\documentclass[12pt]{article}

\usepackage{arxiv}
\usepackage[utf8]{inputenc} 
\usepackage[T1]{fontenc}    
\usepackage{booktabs}       
\usepackage{amsfonts}       
\usepackage{nicefrac}       
\usepackage{microtype}      
\usepackage{lipsum}		
\usepackage{graphicx}
\usepackage{doi}
\usepackage{comment}
\usepackage{amsmath}
\usepackage{mathtools}
\usepackage{multicol}
\usepackage{here}
\usepackage{caption, subcaption}
\usepackage{authblk}
\usepackage{natbib}
\usepackage{url}
\usepackage{hyperref}

\def\widehline{%
\noalign{\global\dimen1 \arrayrulewidth
\global\arrayrulewidth3\arrayrulewidth}%
\hline
\noalign{\global\arrayrulewidth\dimen1}}
\DeclareMathOperator*{\argmax}{argmax} 
\newcommand\rerank{\operatorname{re-rank}}

\title{Predicting from a Different Perspective in Re-ranking Model for Inductive Knowledge Graph Completion}

\author[1]{Yuki Iwamoto}
\author[1]{Ken Kaneiwa}
\affil[1]{
    Department of Computer Network Engineering \\
    Graduate School of Informatics and Engineering\\
    The University of Electro-Communications\\
    Tokyo, Japan
}
\affil{\texttt{iwamoto@sw.cei.uec.ac.jp}, \texttt{kaneiwa@uec.ac.jp}}

\begin{document}
\maketitle              
\begin{abstract}
Rule-induction models have demonstrated great power 
in the inductive setting of knowledge graph completion.
In this setting, 
the models are tested on a knowledge graph entirely composed of unseen entities.
These models learn relation patterns as rules by utilizing subgraphs.
Providing the same inputs with different rules leads to differences in the model's predictions.
In this paper, we focus on the behavior of such models.
We propose a re-ranking-based model called ReDistLP (\textit{Re-ranking with a Distinct Model for Link Prediction}).
This model enhances the effectiveness of re-ranking by 
leveraging the difference in the predictions between the initial retriever and the re-ranker.
ReDistLP outperforms the state-of-the-art methods in 2 out of 3 benchmarks.

\end{abstract}

\section{Introduction}
Knowledge graphs (KGs) store collections of facts in the form of triples.
Each triple $(h, r, t)$ represents a relation $r$ between a head entity $h$ and a tail entity $t$.
Various KGs have been proposed,
including Freebase \citep{bollacker_freebase_2008} for real-world data 
and GeneOntology \citep{gene_gene_2004} for biological data.
These KGs are used in many applications, such as question-answering and recommendation systems; however, since many KGs are inherently incomplete,
completing the missing information is essential.
One way to do this is by link prediction, which is a task that predicts missing links in a KG.
Such Knowledge Graph Completion (KGC) improves the KG itself and enhances the usefulness of applications using the KG

Knowledge Graph Embedding (KGE), such as TransE \citep{bordes2013transe} and RotatE \citep{sun2019rotate}, is the most dominant method for link prediction.
Most KGE methods are designed for the transductive setting,
where the test entities are a subset of the training entities.
These models may not produce meaningful embeddings for unseen entities in the inductive setting.
In the inductive setting, where entities for testing are absent in the training data,
while the relations remain the same\citep{teru_inductive_2020_3}.
This setting requires the model to have a high inductive ability.
Inductive ability is the capability to generalize and make accurate predictions for unseen entities based on the learned patterns and rules from the training data.

GraIL \citep{teru_inductive_2020_3} and BERTRL \citep{zha_inductive_2021_2} are designed for the inductive setting.
They learn the probabilistic logical rules of relation patterns
using subgraphs around an input triple.
These models demonstrate high inductive ability because 
such rules are entity-independent.

Our empirical findings reveal that BERTRL tends to predict different entities when distinct rules are provided as input.
The model trained with the longer rules predicts entities within narrower regions on a KG.
Motivated by this observation, we propose ReDistLP (\textit{Re-ranking with a Distinct Model for Link Prediction}), which leverages this phenomenon to improve accuracy.
ReDistLP introduces a re-ranking pipeline using two variants of BERTRL that are trained with different lengths of rules.
The re-ranking pipeline obtains a pool of candidate documents through an inexpensive approach and subsequently re-evaluates them using a more expensive model \citep{wang2011cascade-model}.
This approach achieves both low computational cost and high accuracy.
While existing research has mainly focused on improving the recall of the initial retriever or the performance of the re-ranker, 
this study focuses on the difference in the predictions between them.
We show theoretical insights to improve the accuracy of the re-ranking pipeline.
When selecting a model as the re-ranker,
choosing one that predicts different entities compared to the initial retriever can improve accuracy. 
We train three variants of BERTRL,
each incorporating rules of different lengths as additional inputs during training and inference.
We construct ReDistLP using these variants of BERTRL.
ReDistLP achieves the best performance in 2 out of 3 benchmarks. 

\section{Related Works and Backgrounds}
\label{Related Works and Backgrounds}

\subsection{Knowledge Graph Completion Models}

\subsubsection{Rule-induction Methods.}
Rule-induction methods learn probabilistic logical rules
from a KG.
This method inherently can solve the inductive setting
since such rules are independent of entities.
Here is an example of a rule
\begin{eqnarray}
(x, {\rm president\_of}, y) \land (z, {\rm capital\_of}, y) \to (x, {\rm work\_at}, z).
\end{eqnarray}
This rule denotes that 
if $x$ is the president of $y$, and $z$ is the capital of $y$, then $x$ works at $z$.
Once such rules are learned, they can be used to generalize unseen entities.
AMIE \citep{galarraga2013amie} and RuleN \citep{meilicke2018rulen} mine these rules while reducing the search space.
NeuralLP \citep{yang2017neurallp} and DRUM \citep{sadeghian2019drum}
learn such rules in an end-to-end differentiable manner.
GraIL \citep{teru_inductive_2020_3} uses a graph neural network to learn these rules over the subgraphs around the input triple.

\subsubsection{Language-based Models.}
KG-BERT \citep{yao2019kg-bert} and BERTRL \citep{zha_inductive_2021_2}
are language-based models that use pre-trained language models to 
utilize the textual information.
KG-BERT takes the triple as textual sequences.
It trains BERT to score the plausibility of the input triple.
KG-BERT has limited inductive ability since they do not utilize subgraph information.
Their inductive capacity is solely attributed to the pre-trained language model.
In contrast, BERTRL demonstrates high inductive ability.
Similar to KG-BERT,
BERTRL is a model designed to identify triples.
This model takes the paths connecting head and tail entities and the input triple.
Its high inductive capability arises from 
both a pre-trained language model 
and a subgraph reasoning mechanism.
Hence, BERTRL can also be classified as a subgraph-induction method.
The formulation for BERTRL is as follows
\begin{equation}
score(h, r, t) = \max_{\rho} p(y=1|h,r,t, \rho),
\label{bertrl}
\end{equation}
where $\rho$ denotes a path connecting entities $h$ and $t$.
The score of an input triple is the maximum score obtained over all possible paths.

\section{Analysis of Model Behavior Under Different Rule Sets}
\label{Analyzing Model Behavior Under Different Rule Sets}
This section demonstrates that BERTRL predicts distinct entities when
different rules are provided as input.
We train BERTRL with 3-hop, 2-hop, and 1-hop rules, respectively, and analyze the differences in their prediction sets.

We investigate the dissimilarities in predictions between variant models from the intersection ratio of top-$n$ prediction sets in Table \ref{analyzing intersection ratio}.
The results show that the prediction sets significantly depend on the number of hops used in the training rules.
The predictions of BERTRL trained with 1-hop and 2-hop rules are the most similar, while those trained with 1-hop and 3-hop rules are the least similar.

\noindent
\begin{minipage}[b]{.4\hsize}
    \small
    \centering
    \captionof{table}{Intersection ratio of top-$n$ prediction sets}
    \label{analyzing intersection ratio}
    \begin{tabular}{cccc}
    \widehline
    & WN18RR & FB15k-237 & NELL-995 \\
    \hline\hline
    \multicolumn{4}{c}{$n=10$} \\
    1-hop 2-hop & 0.0711 & 0.372 & 0.471\\
    1-hop 3-hop & 0.0364 & 0.274 & 0.354 \\
    2-hop 3-hop & 0.0549 & 0.342 & 0.430 \\
    \hline
    \multicolumn{4}{c}{$n=100$} \\
    1-hop 2-hop & 0.236 & 0.431 & 0.606 \\ 
    1-hop 3-hop & 0.104 & 0.282 & 0.476 \\
    2-hop 3-hop & 0.103 & 0.294 & 0.514 \\
    \hline
    \multicolumn{4}{c}{$n=1000$} \\
    1-hop 2-hop & 0.921 & 0.970 & 0.790 \\ 
    1-hop 3-hop & 0.921 & 0.970 & 0.790 \\
    2-hop 3-hop & 0.921 & 0.970 & 0.790 \\
    \widehline
    \end{tabular}
\end{minipage}
\hspace{4.5em}
\begin{minipage}[b]{.4\hsize}
    \small
    \centering
    \captionof{table}{$k$-hop reachability between entities in top-10 predicted sets}
    \label{analyzing reachability table}
    \begin{tabular}{cccc}
        \widehline
        & WN18RR & FB15k-237 & NELL-995 \\
        \hline
        \hline
        \multicolumn{4}{c}{$k=2$} \\
        3-hop & 0.338 & 0.639 & 0.894 \\
        2-hop & 0.178 & 0.612 & 0.899 \\
        1-hop & 0.0624 & 0.283 & 0.538 \\
        \hline
        \multicolumn{4}{c}{$k=3$} \\
        3-hop & 0.441 & 0.860 & 0.980 \\
        2-hop & 0.239 & 0.777 & 0.958 \\
        1-hop & 0.110 & 0.487 & 0.740 \\
        \hline
        \multicolumn{4}{c}{$k=5$} \\
        3-hop & 0.577 & 0.958 & 0.996 \\
        2-hop & 0.392 & 0.910 & 0.992 \\
        1-hop & 0.282 & 0.797 & 0.959 \\
        \widehline
    \end{tabular}
\end{minipage}
\vspace{2em}

\begin{figure}[h]
\centering
\begin{subfigure}[b]{.3\textwidth}
    \includegraphics[width=1.1\linewidth]{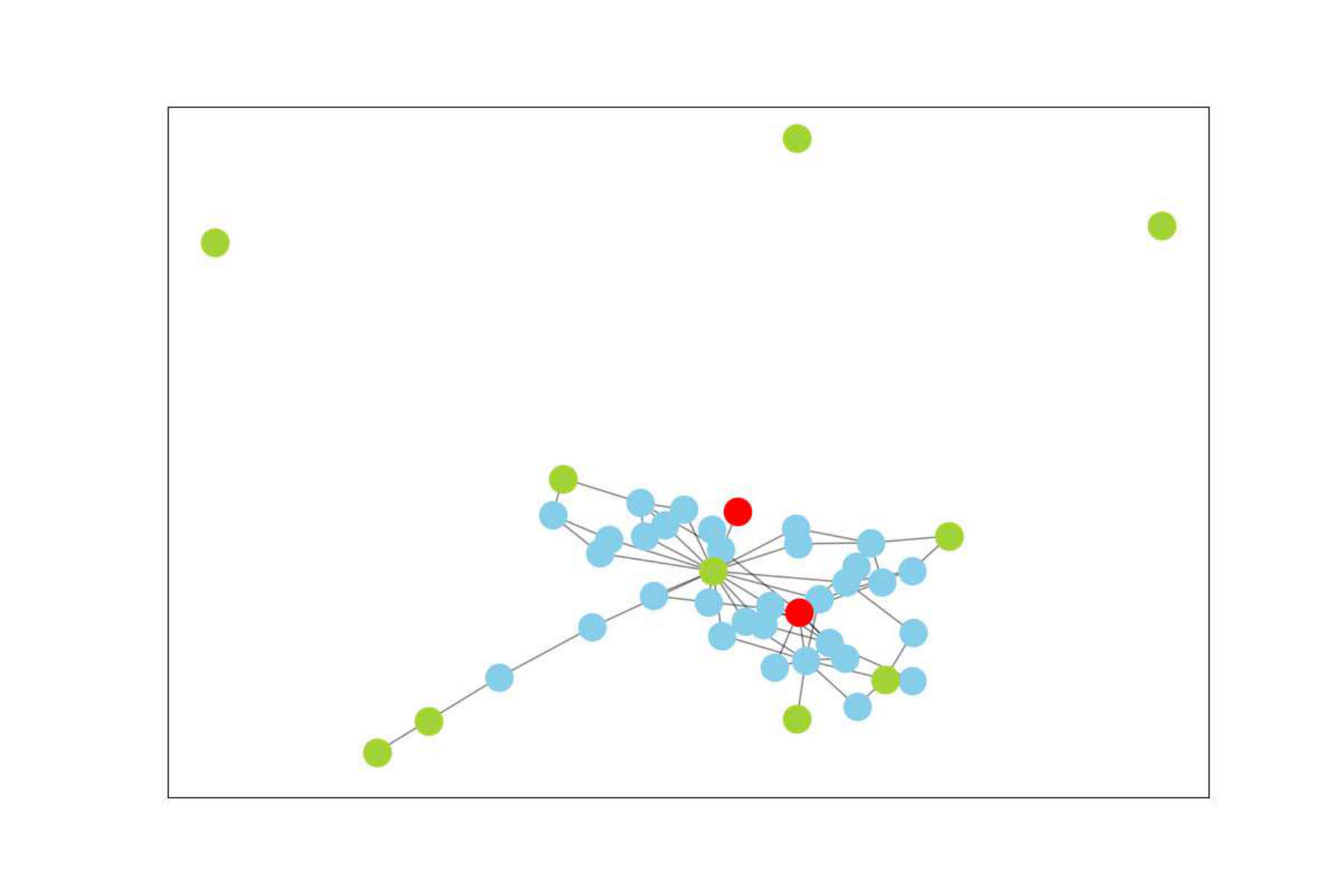}
    \caption{1-Hop (FB15k-237)}
\end{subfigure}
\begin{subfigure}[b]{.3\textwidth}
    \includegraphics[width=1.1\linewidth]{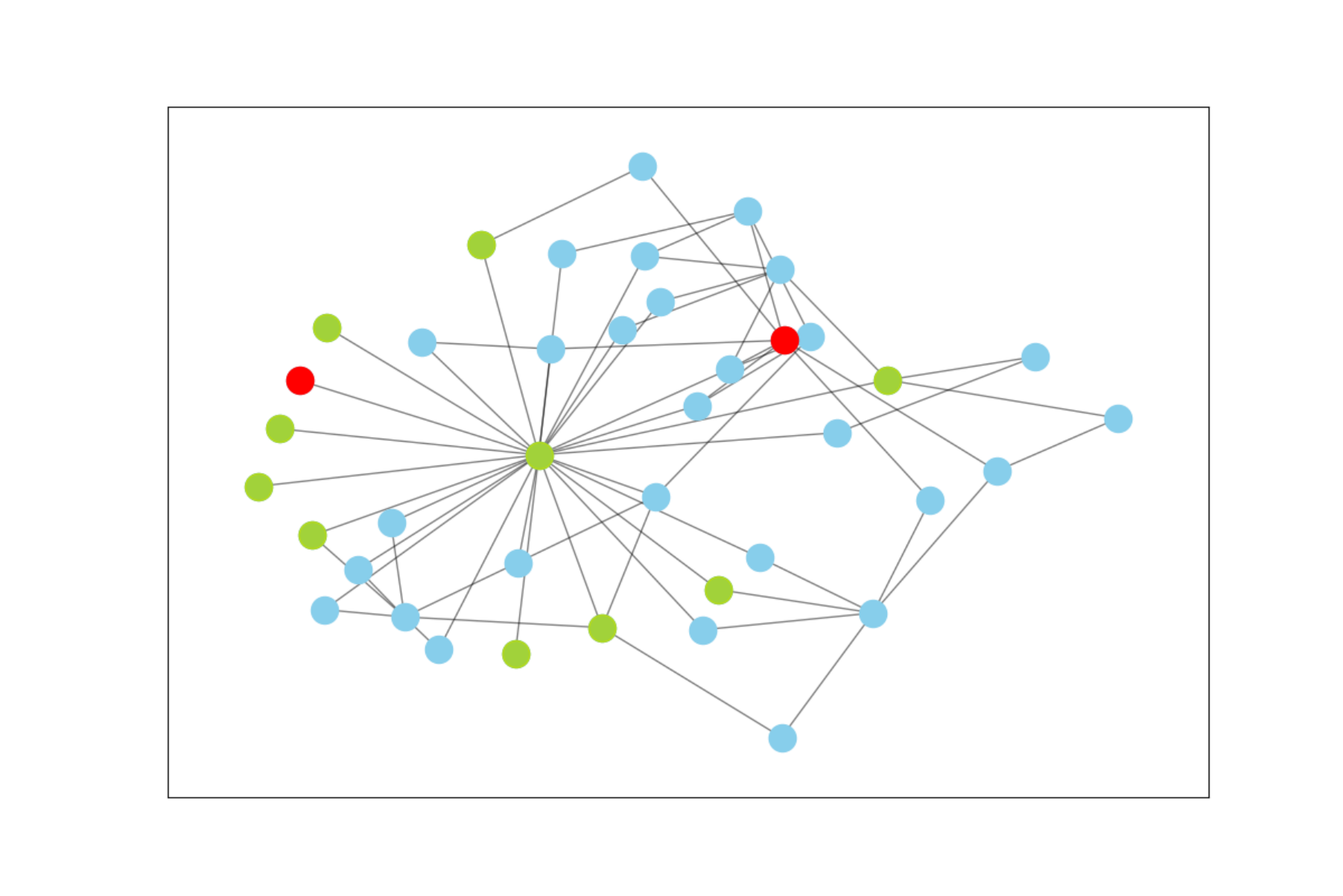}
    \caption{2-Hop (FB15k-237)}
\end{subfigure}
\begin{subfigure}[b]{.3\textwidth}
    \includegraphics[width=1.1\linewidth]{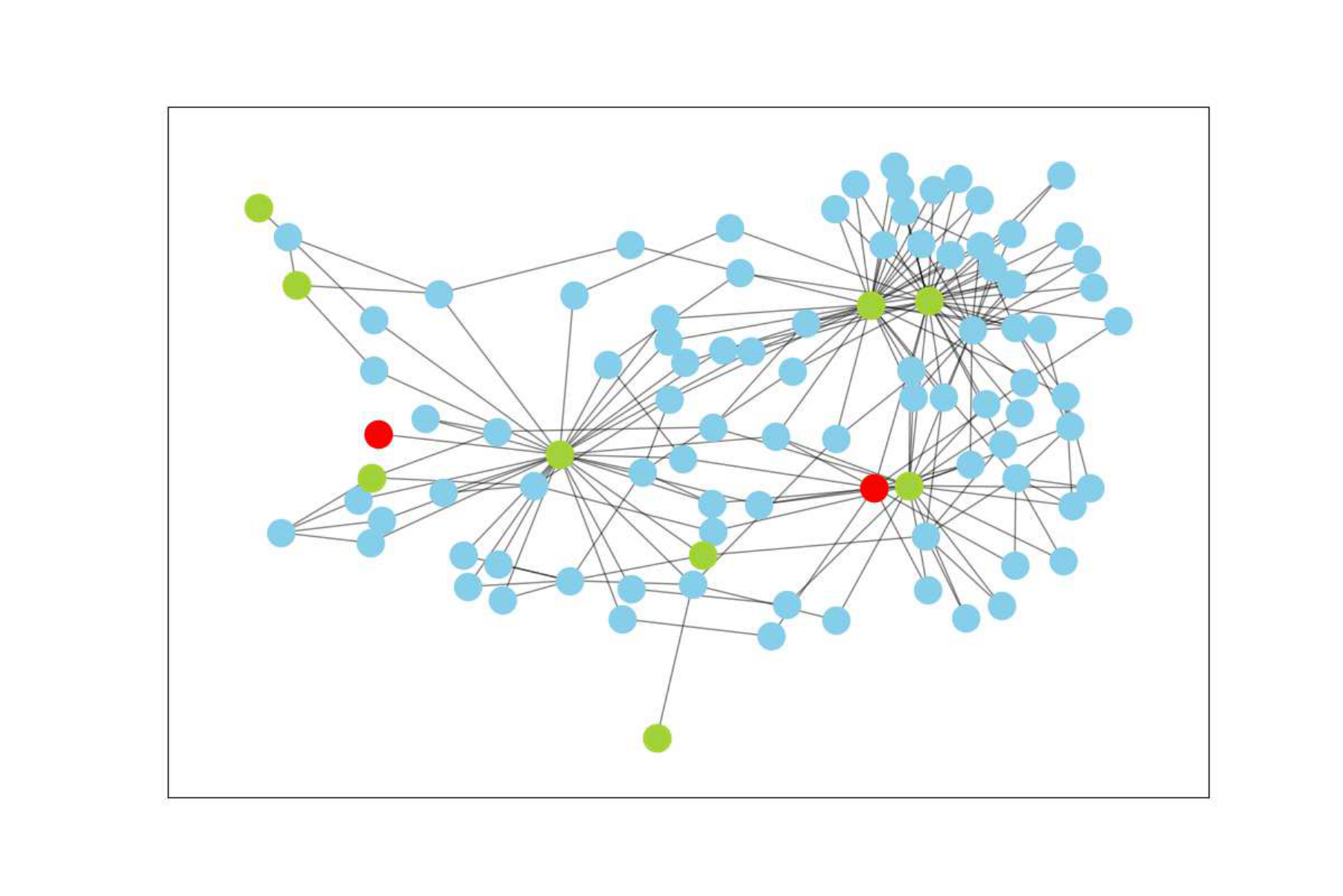}
    \caption{3-Hop (FB15k-237)}
\end{subfigure}
\caption{Visualization of predicted top-10 entities. Red nodes represent entities from the input triple. Green nodes represent predicted entities. Blue nodes represent entities that appear in connected paths.}
\label{analyzing reachability fig1}
\end{figure}

Furthermore, we observe that the longer the number of hops, the more the predicted entities tend to be concentrated locally on the KG.
Figure \ref{analyzing reachability fig1} shows 
the top-10 predicted entities and the paths that connect them within 3-hop.
Table \ref{analyzing reachability table} shows 
$n$-hop reachability between entities in top-10 predicted sets.

In Table \ref{analyzing reachability table} and Figure \ref{analyzing reachability fig1}, it can be observed that:
(1) Reachability in predicted entities increases with the number of hops of rules.
(2) BERTRL with a higher number of hops predicts entities within narrower regions on a KG, while predicted entities by BERTRL with fewer hops spread more widely on a KG.

\section{Proposed Approach}
\label{Proposed Approach}
This section first provides insight into enhancing re-ranking effectiveness for our model.
Then, we introduce our model of ReDistLP and explain how to re-rank the candidate entities for our model.

\subsection{Maximize the Effectiveness of Re-ranking}
\label{Maximize the Effectiveness of Re-ranking}

We provide theoretical insights to maximize re-ranking effectiveness through fuzzy sets.
The predictions of a model $m$ can be noted using fuzzy sets.
\begin{equation}
    S_{m} = \{ (t, \mu_{S_{m}}(t)) | t \in \mathcal{E} \},
\end{equation}
where $\mu_{S_{m}}$ is a membership function
that assigns to item membership degree between 0 and 1 \citep{zimmermann_fuzzy_2010}.
We use the model predictions as the membership function, i.e., $\mu_{S_{m}}(t) = score_{m}(h, r, t)$.

Let $\bar{t} \in \mathcal{E}$ denote the correct answer for the input triple $(h, r, ?)$.
Now, we consider $\alpha$-cut for the fuzzy set.
The $\alpha$-cut of a fuzzy set is a classical set containing all the elements with a membership degree greater than or equal to a value $\alpha$ \citep{zimmermann_fuzzy_2010}.
\begin{equation}
S_{m}^{\geq \mu_{S_m}(\bar{t})} = \{t \in \mathcal{E} |
\mu_{S_m}(t) \geq \mu_{S_m}(\bar{t})\}.
\end{equation}
The size of $S_{m}^{\geq \mu_{m}(\bar{t})}$ is equivalent to the performance of model $m$ for query $(h,r, ?)$,
because $1/|S_{m}^{\geq \mu_{m}(\bar{t})}|$ corresponds to the Reciprocal Rank (RR).
The Mean Reciprocal Rank (MRR) is widely used 
as an evaluation metric
in research fields such as information retrieval systems, recommender systems, and KGC.

Suppose that model $m_{1}$ is a initial retriever,
and model $m_{2}$ is a re-ranker.
Let us assume the accuracy of $m_{2}$ exceeds that of $m_{1}$ (
$|S^{\geq \mu_{S_{m_1}}(\bar{t})}_{m_{1}}| > |S^{\geq \mu_{S_{m_2}}(\bar{t})}_{m_{2}}|$).
After re-ranking, the set of predicted entities is given by
\begin{equation}
S^{\prime}_{m_1, m_2} = \{t \in S_{m_1}^{\geq \mu_{S_{m_1}}(\bar{t})}|
    \mu_{S_{m_{2}}}(t) \geq \mu_{S_{m_{2}}}(\bar{t}) \}. 
\label{s prime m1 m2}
\end{equation}
Given the above, we can infer
\begin{equation}
\bigl|S^{\prime}_{m_1,m_{2}}\bigr| \leq \bigl|S_{m_{2}}^{\geq \mu_{S_{m_2}}(\bar{t})} \bigr|
< \bigl|S_{S_{m_{1}}}^{\geq \mu_{S_{m_1}}(\bar{t})}\bigr|.
\label{sm1 <= sm2}
\end{equation}
This equation indicates that the performance after re-ranking exceeds that of the initial retriever and the re-ranker.
Equation (\ref{sm1 <= sm2}) can be resolved into the following equation and inequality,
depending on the relationship between $S_{m_{1}}^{\geq \mu_{S_{m_1}}(\bar{t})}$ and $S_{m_{2}}^{\geq \mu_{S_{m_2}}(\bar{t})}$:
\begin{equation}
S_{m_{1}}^{\geq \mu_{S_{m_1}}(\bar{t})} \subset 
S_{m_{2}}^{\geq \mu_{S_{m_2}}(\bar{t})} \implies
\bigl|S^{\prime}_{m_1,m_2}\bigr| = \bigl|S_{m_2}^{\geq \mu_{S_{m_2}}(\bar{t})}\bigr|
\end{equation}
This equation suggests that when $S^{\geq \mu_{S_{m_1}}(\bar{t})}_{m_{1}}$ is a subset of $S^{\geq \mu_{S_{m_2}}(\bar{t})}_{m_{2}}$,
there will be no improvement in accuracy after re-ranking.
On the other hand, if $S_{m_{1}}^{\geq \mu_{S_{m_1}}(\bar{t})}$ is not a subset of $S_{m_{2}}^{\geq \mu_{S_{m_2}}(\bar{t})}$,
the following equation holds, implying that the accuracy after re-ranking will be higher than that of the re-ranker alone:
\begin{equation}
S_{m_{1}}^{\geq \mu_{S_{m_1}}(\bar{t})} \not\subset 
S_{m_{2}}^{\geq \mu_{S_{m_2}}(\bar{t})} \implies 
\bigl|S^{\prime}_{m_1,m_2}\bigr| < \bigl|S_{m_2}^{\geq \mu_{S_{m_2}}(\bar{t})}\bigr|
    \label{s m1 not subset s m2}
\end{equation}

Furthermore,
by selecting models based on the next equation,
re-ranking effectiveness is maximized through the minimization of $S^{\prime}_{m_{1},m_{2}}$:
\begin{equation}
    \argmax_{\;\;m_{1}, m_{2}} \; \Bigl|S_{m_1}^{\geq \mu_{S_{m_1}}(\bar{t})} \setminus S_{m_2}^{\geq \mu_{S_{m_2}}(\bar{t})}\Bigr|
    \label{S m1 setminus S m2}
\end{equation}

From the above observations, 
it is ineffective to select the models that make almost the same predictions;
it is crucial to choose the models that offer distinct predictions.

However, in practical use,
no guarantee that the correct answer
$\bar{t}$ is in the initial pool.
This is because the initial retriever selects from top-$k$ instances.
Therefore, 
it is important to maximize the effectiveness of re-ranking,
considering both the performance of the initial retriever $m_{1}$ and the relationship between models
in practical use.

\subsection{Model Details}
\label{Model details}
\begin{figure}[t]
\centering
\includegraphics[width=\linewidth]{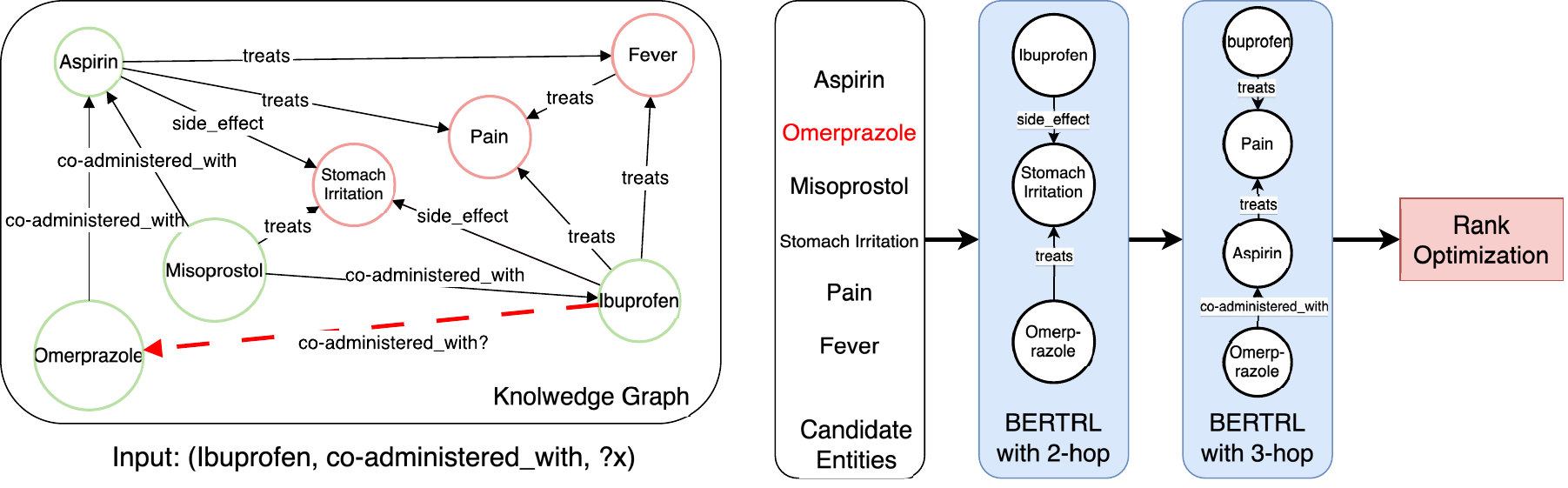}
\caption{An overview of ReDistLP.}
\label{abst2}
\end{figure}

Figure \ref{abst2} provides an overview of our ReDistLP model.
ReDistLP is a cascaded model that first scores candidate entities using an initial retriever and then re-ranks them with a re-ranker.
As discussed in Section \ref{Maximize the Effectiveness of Re-ranking}, we aim to select the initial retriever and re-ranker models such that their prediction sets ($S_{m_{1}}$ and $S_{m_{2}}$) satisfy the condition ($S_{m_{1}} \not\subset S_{m_{2}}$),
according to equation (\ref{s m1 not subset s m2}).
This non-subset relationship between the prediction sets can enhance the effectiveness of the re-ranking process.
As shown in the analysis in Section \ref{Analyzing Model Behavior Under Different Rule Sets}, when trained with longer hop rules, the BERTRL tends to predict entities concentrated within narrower regions of the KG.
To satisfy the condition in equation (\ref{s m1 not subset s m2}) and maximize re-ranking effectiveness, we employ a BERTRL trained on shorter hop rules as the initial retriever and a BERTRL trained on longer hop rules as the re-ranker
Therefore, in our final ReDistLP, we use BERTRL trained with 2-hop as the initial retriever and BERTRL trained with 3-hop as the re-ranker.

The initial retriever scores candidate entities $C$ using BERTRL trained with 2-hop rules.
The prediction of the initial retriever is as follows:
\begin{equation}
S_{I}(h, r) = \bigl\{ ((h, r, t), score_{{\text{BERTRL 2-hop}}}(h, r, t)) | t \in C \bigr\}
\label{S_i definition}
\end{equation}
The re-ranker scores the candidate entities using BERTRL trained with 3-hop rules, denoted as $S_{R}(h, r)$, which follows a similar definition to equation (\ref{S_i definition}).
The re-ranked prediction set $S^{\prime}(h,r)$ is obtained from the initial prediction set $S_{I}(h,r)$ and the scores $S_{R}(h, r)$,
where the final score for each entity $e$ is given by $\mu_{S^{\prime}(h,r)}(e)$.

\subsection{Rank Threshold Optimization}
\label{Rank Threshold Optimization}
The optimal cutoff point for the initial retriever's results directly affects the accuracy of the re-ranking model.
Ideally, the cutoff would be defined as the position of the answer entity.
However, in practice, 
model predictions are continuous probability scores,
which can be noisy, making it difficult to identify a stable threshold.
We explored different threshold optimization methods to find the most suitable approach for our model.

\subsubsection{Classical Sets Approach.}
Classical sets, also known as crisp sets, have well-defined boundaries where an element either belongs to the set or does not.
In the context of re-ranking, we can apply classical set theory to determine where to truncate the ranking list generated by the initial retriever.

\paragraph{Threshold.}
One straightforward approach is to use a constant threshold,
where a fixed threshold is applied to the ranking list.
The re-ranking function using a threshold $\theta$ is defined as
\begin{equation}
    \rerank_{\theta}(S_{I}, S_{R}) =
    \Bigl\{
    \Bigl( (h,r,t), \mu_{S_{R}}(t)\Bigr) | t \in S_{I}(h,r)^{\geq \theta}
    \Bigr\}.
\end{equation}
The re-ranking function selects the triples from the $\theta$-cut of the initial prediction set.
We determine the optimal fixed threshold $\theta$ using a development dataset and then apply it to the test set.

\paragraph{Top-$k$ Cutoff.}
Another common approach is to use a top-k cutoff, where the top-$k$ items from the ranked list are selected.
This method ensures a fixed number of items are always selected, regardless of their absolute scores.
The re-ranking function using a top-$k$ cutoff is given by:
\begin{equation}
    \rerank_{{\text top}\mbox{-}k}(S_{I}, S_{R}) =
    \Bigl\{
    \Bigl( (h,r,t), \mu_{S_{R}}(t)\Bigr) | t \in {\text top}\mbox{-}k(S_{I}(h,r))
    \Bigr\}
\end{equation}
where the function ${\text top}\mbox{-}k(S_{I}(h,r))$ selects the top-$k$ entities from the initial prediction set based on their membership scores.

\paragraph{$k$-means clustering.}
A third approach is to use k-means clustering, where the items in the ranked list are grouped into $k$ clusters based on their similarity.
Items belonging to the top-scoring cluster(s) are selected for re-ranking.
This method groups similar items together and selects the most relevant cluster(s) based on their overall scores.
The re-ranking function using $k$-means clustering can be expressed as:
\begin{equation}
    \rerank_{k\mbox{-}{\text means}}(S_{I}, S_{R}) =
    \Bigl\{
    \Bigl( (h,r,t), \mu_{S_{R}}(t)\Bigr) | (h, r, t) \in \bigcup_{j=1}^{m} C_{j}
    \Bigr\}
\end{equation}
where 
the function $\bigcup_{j=1}^{m} C_{j}$ represents the union of the top-$m$ clusters obtained from applying $k$-means clustering to the probability distribution of the initial prediction set $S_{I}(h,r)$.

\subsubsection{Fuzzy Sets Approach.}
The classical sets approach has a major drawback: it often excludes the answer entity from the selected set.
When the answer entity is not included in the initial pool, the re-ranking model's accuracy is limited to that of the initial retriever.
In contrast, the fuzzy sets approach can be employed to mitigate this issue.
\paragraph{Intersection.}
The equation (\ref{s prime m1 m2}) is equal to 
$S_{m_1}^{\geq \mu_{m_{1}}(\bar{t})} \cap S_{m_2}^{\geq \mu_{m_{2}}}(\bar{t})$.
Generalizing this approach leads us to the definition of the intersection-based re-ranking function: 
$\rerank_{\cap}(S_{I}, S_{R}) = S_{I} \cap S_{R}$.
The membership function of the resulting fuzzy set is given by:
\begin{equation}
    \mu_{\rerank_{\cap}(S_i, S_r)}(t) = 
    {\text min}(\mu_{S_{I}}(t), \mu_{S_{R}}(t))
\end{equation}

\paragraph{Union.}
In fuzzy set theory,
the union of two fuzzy sets can be defined using the maximum operator \citep{zadeh_fuzzy_1965}.
Applying this concept, we introduce the union-based re-ranking function $\rerank_{\cup}(S_{I}, S_{R})$.
The membership function of the resulting fuzzy set for the union-based re-ranking approach is given by:
\begin{equation}
    \mu_{\rerank_{\cup}(S_i, S_r)}(t) = 
    {\text max}(\mu_{S_{I}}(t), \mu_{S_{R}}(t))
\end{equation}
\paragraph{Convex combination (Mean).}
Another approach to combining fuzzy sets is through a convex combination.
The convex combination of $A$, $B$, and $\Lambda$ is represented by 
$(A, B; \lambda)$ \citep{zadeh_fuzzy_1965}.
The intersection, convex combination, and union of fuzzy sets satisfy the following relationship \citep{zadeh_fuzzy_1965}:
\begin{equation}
   A \cap B \subset (A, B; \Lambda) \subset A \cup B
    \quad\text{for all $\Lambda$}
\end{equation}
This relationship implies that the convex combination of two fuzzy sets is always bounded by their intersection and union,
regardless of the weight fuzzy set $\Lambda$ choice.
The convex combination has a softened membership function compared with the intersection.
Hence, we introduce the mean-based re-ranking function $\rerank_{\text{mean}}(S_{I}, S_{R})$ as follows:
\begin{equation}
   \rerank_{{\text mean}} = (S_{I}, S_{R}; \Lambda)
\end{equation}
where $\Lambda$ is a fuzzy set representing the weights of the convex combination.
We define $\Lambda$ as a fuzzy set with a membership value of 0.5 for all entities.
The membership function of the resulting fuzzy set is given by:
\begin{equation}
    \mu_{(S_{I}, S_{R}; \Lambda)}(t) = 
    \mu_{\Lambda}(t) \mu_{S_{I}}(t) + (1 - \mu_{\Lambda}(t)) \mu_{S_{R}}(t)
\end{equation}
Our final model uses this re-ranking function.

\subsection{Training}
We make two variants of datasets for BERTRL in training and inference.
As the original one used in the previous study,
the first variant of the dataset excludes the triples where the head and tail entities do not connect within $n$-hop.
The second variant does not exclude such triple data.
Table \ref{training data} shows examples of training data.
In this case, the first dataset only contains No.1 and 2,
while the second dataset contains all cases.
This enables the model to score entities that do not have any connecting path to the other entity in a given triple within the n-hop.
We call BERTRL tested on the second dataset as BERTRL v2,
which is used as the initial retriever for our model.

\begin{table}[h]
    \caption{Examples of training data from NELL-995}
    \label{training data}
    \centering
    \begin{tabular}{cp{10cm}}
    \widehline
     No.1 & Question: country : republic country also known as what ? Is the correct answer country : vietnam ?   Context: country : republic country located in geopolitical location state or province : states; state or province : states state located in geopolitical location country : vietnam; \\
     No.2 &  Question: country : republic country also known as what ? Is the correct answer country : vietnam ?   Context: country : republic country located in geopolitical location state or province : states; country : vietnam country located in geopolitical location state or province : states; \\
    No.3 & Question: country : republic country also known as what ? Is the correct answer hospital : hialeah ?  Context: None \\
    No.4 & Question: country : republic country also known as what ? Is the correct answer city : beverly hills ?        Context: None \\
    \widehline
    \end{tabular}
\end{table}
\section{Experiments}
\label{Experiments}

In our experiments,
we evaluate the effectiveness of our proposed model 
on inductive knowledge graph completion datasets.
In the latter part, 
we analyze the performance of our model in the ideal setting.

\subsection{Experiment Settings}
We use inductive subsets of three knowledge completion datasets:
WN18RR \citep{dettmers2018conve}, 
FB15k-237 \citep{toutanova2015representing}, and NELL-995 \citep{xiong2017deeppath},
which were introduced by Teru et al. \citep{teru_inductive_2020_3}.

We report the evaluation metrics Hits@1 and MRR.
Hits@1 measures the percentage of cases where the correct triple is ranked as the top 1 prediction.
MRR computes the average of the reciprocal ranks of the correct triple.
Consistent with the evaluation setup in GRAIL \citep{teru_inductive_2020_3} and other prior work, we measure the rank of the correct triple among
a set of 50 candidate triples, comprising the correct triple and 49 negative triples.

We compare our approach with GraIL \citep{teru_inductive_2020_3},
BERTRL \citep{zha_inductive_2021_2},
KG-BERT \citep{yao2019kg-bert},
RuleN \citep{meilicke2018rulen},
and KRST \citep{zhixiang2023krst}.
We use the implementation provided publicly by the authors and the best hyper-parameter settings reported in their work.



\subsection{Results and Analysis}
\label{Results and Analysis}
\textbf{Actual Setting.}
Table \ref{Hits@1 and MRR in actual setting} presents the Hits@1 and MRR results of inductive link prediction.

\begin{table}[h]
    \centering
    \small
    \caption{Hits@1 and MRR of link prediction in the inductive setting.}
    \label{Hits@1 and MRR in actual setting}
    \begin{tabular}{lrrrrrr}
    \widehline
    & \multicolumn{2}{c}{WN18RR} & \multicolumn{2}{c}{FB15k-237} & \multicolumn{2}{c}{NELL-995} \\
    & Hits@1 & MRR & Hits@1 & MRR & Hits@1 & MRR \\
    \hline\hline
    RuleN & .745 & .780 & .415 & .462 & .638 & .710 \\
    GRAIL & .769 & .799 & .390 & .469 & .554 & .675 \\
    KG-BERT & .436 & .542 & .341 & .500 & .244 & .419 \\
    BERTRL & .755 & .792 & .541 & .605 & .715 & .808 \\
    KRST & {\bfseries .835} & {\bfseries .902} & .602 & {\bfseries .716} & .649 & .769 \\
    \hline
    BERTRL (1 hop) & .630 & .661 & .329 & .390 & .466 & .519 \\
    BERTRL (2 hop) & .729 & .749 & .410 & .455 & .639 & .698 \\
    BERTRL (3 hop) & .793 & .815 & .578 & .636 & .731 & .821 \\
    

    \hline
    ReDistLP (2 to 3-hop) threshold $\theta$ & .794 & - & .627 & - &  .735 & - \\
    ReDistLP (2 to 3-hop) top-$k$ cutoff ($k=40$)& .780 & - & .607 & - & .736 & - \\
    ReDistLP (2 to 3-hop) $k$-means (3 out of 5) & .775 & - &  .620 & - & .719 & - \\
    ReDistLP (2 to 3-hop) Intersection & .794 & .814 &.600 & .631 & .721 & .805 \\
    ReDistLP (2 to 3-hop) Union & .761 & .793 & .627 & .706 & .746 & .831 \\
    ReDistLP (2 to 3-hop) Mean & .796 & .808 & {\bfseries .634} & .715 & {\bfseries .748} & {\bfseries .832} \\
    \widehline
    \end{tabular}
\end{table}
Our model achieves the best performance under Hits@1 on the FB15k-237 and NELL-995 datasets.
Specifically, ReDistLP achieves significant improvements in the FB15k-237
(+3.2\% for Hits@1) and NELL-995 (+3.3\% for Hits@1).
Under the MRR metric, our model achieves the best performance on the NELL-995 dataset and comparable results on the FB15k-237 dataset.
In WN18RR, however, KRST obtains the highest. 
Although our model does not achieve the best performance on WN18RR, it still outperforms BERTRL under the Hits@1 and MRR metrics.
ReDistLP, which combines a 2-hop retriever and a 3-hop re-ranker,
outperforms BERTRL (3-hop) on all datasets for both Hits@1 and MRR metrics, 
demonstrating the effectiveness of our redistribution approach.

\textbf{Ideal Setting.}
Table \ref{Hits@1 and MRR in the ideal setting} presents the results of inductive link prediction in the ideal setting.
In this ideal setting, we determine the cutoff based on the position of the answer entity in the initial pool.
\begin{table}[h]
    \centering
    \caption{Hits@1 and MRR results on the ideal setting}
    \label{Hits@1 and MRR in the ideal setting}
    \begin{tabular}{lrrrrrrc}
    \widehline
    & \multicolumn{2}{c}{WN18RR} & \multicolumn{2}{c}{FB15k-237} & \multicolumn{2}{c}{NELL-995} \\
    & H@1 & MRR & H@1 & MRR & H@1 & MRR & Avg. MRR\\
    \hline\hline
    ReDistLP (1 to 2-hop) top-$k$ & .725 & .742 & .590 & .677 & .745 & .818 & .746 \\
    ReDistLP (1 to 3-hop) top-$k$ & .784 & .806 & .720 & .781 & .827 & .891 & .826 \\
    ReDistLP (2 to 3-hop) top-$k$ & .817 & .836 & .705 & .777 & .831 & .893 & {\bfseries .835} \\
    \hline
    ReDistLP (2 to 3-hop) {\scriptsize Intersection} & .820 & .850 & .695 & .781 & .742 & .827 & .819 \\
    ReDistLP (2 to 3-hop) Union & .775 & .797 & .671 & .738 &  .805 & .872 & .802 \\
    ReDistLP (2 to 3-hop) Mean & .801 & .821 & .661 & .736 & .772 & .850 & .802 \\
    \widehline
    \end{tabular}
\end{table}

The Hits@1 metrics show a large difference between the actual setting and the ideal setting.
Specifically, the Hits@1 of ReDistLP (2 to 3-hop) in the ideal setting is higher than in the actual setting by 2.6\%, 7.1\%, and 8.3\% on WN18RR, FB15k-237, and NELL-995, respectively.
This indicates if we could determine a more optimal cutoff,
our model's accuracy could be improved.

ReDistLP (2 to 3-hop) shows the best results.
However, when compared with the initial retriever of BERTRL (Table \ref{Hits@1 and MRR in actual setting}),
ReDistLP (1 to 3-hop) demonstrates the largest improvements.
To better understand the factors contributing to these differences in performance,
we investigate the difference between their prediction sets.
We calculate the following equation
\begin{equation}
    r = \frac{|S_{I} \setminus S_{R}|}{|S_{I}|},
    \label{mean set diff}
\end{equation}
where $S_{I}$ and $S_{R}$ represent the prediction sets of the initial retriever and the re-ranker, respectively.
This equation, which corresponds to equation (\ref{S m1 setminus S m2}),
normalizes the difference in cardinality between the two prediction sets.
A higher value of $r$ indicates a larger difference between the two sets.
Table \ref{distinctiveness of predictions} shows the mean of $r$ calculated over the test set.

\begin{table}[h]
\centering
\caption{Analysis of the difference in prediction sets between the initial retriever and re-ranker.} 
\label{distinctiveness of predictions}
\begin{tabular}{llccc}
    \widehline
    & & WN18RR & FB15k-237 & NELL-995 \\
    \hline
    \hline
    BERTRL 1-hop & BERTRL 2-hop & 0.192 & 0.222 & 0.190 \\
    BERTRL 1-hop & BERTRL 3-hop & 0.234 & 0.259 & 0.237 \\
    BERTRL 2-hop & BERTRL 3-hop & 0.192 & 0.221 & 0.191 \\
    \widehline
\end{tabular}
\end{table}

Table \ref{distinctiveness of predictions} reveals that 
the pair of (BERTRL 1-hop, BERTRL 3-hop) consistently yields the highest values.
This suggests that improving accuracy is achieved by 
maximizing the difference in prediction sets between the initial retriever and re-ranker.

Compared to ReDistLP variants, the classical set approaches, such as $k$-means, top-$k$, and fixed threshold, could not outperform the fuzzy set approach in the actual setting.
In the ideal setting, however, the classical approach achieves better Hits@1 and Avg. MRR results than all of the fuzzy set approaches.
This implies that determining the optimal cutoff in the initial retriever's results is challenging in our model.
The set must be minimized while still ensuring that the correct entity is included.
We will address this issue in future studies to improve its performance further.

\section{Conclusion}
\label{Conclusion}
We proposed ReDistLP, a re-ranking model designed for inductive link prediction on KGs.
ReDistLP employs two variants of BERTRL, which are trained with different rules.
We analyzed that different rules lead to changes in the model's behavior.
Specifically, the model trained with longer hop rules predicts entities within narrower regions of the KG. 
ReDistLP leverages this phenomenon through a re-ranking pipeline to improve link prediction performance.
Our model outperforms the baselines in 2 out of 3 datasets.

Furthermore, we showed that our model's performance can be further improved if an optimal rank threshold for re-ranking could be determined.
In the future, we plan to optimize the rank threshold for our model.

\bibliography{custom2}

\begin{thebibliography}{18}
\providecommand{\natexlab}[1]{#1}
\providecommand{\url}[1]{\texttt{#1}}
\expandafter\ifx\csname urlstyle\endcsname\relax
  \providecommand{\doi}[1]{doi: #1}\else
  \providecommand{\doi}{doi: \begingroup \urlstyle{rm}\Url}\fi

\bibitem[Bollacker et~al.()Bollacker, Evans, Paritosh, Sturge, and Taylor]{bollacker_freebase_2008}
Kurt~D. Bollacker, Colin Evans, Praveen~K. Paritosh, Tim Sturge, and Jamie Taylor.
\newblock Freebase: a collaboratively created graph database for structuring human knowledge.
\newblock In \emph{{SIGMOD} 2008}, pages 1247--1250. {ACM}.
\newblock URL \url{https://doi.org/10.1145/1376616.1376746}.

\bibitem[Consortium(2004)]{gene_gene_2004}
Gene~Ontology Consortium.
\newblock The gene ontology (go) database and informatics resource.
\newblock \emph{Nucleic acids research}, 32\penalty0 (suppl\_1):\penalty0 D258--D261, 2004.

\bibitem[Bordes et~al.()Bordes, Usunier, Garc{\'{\i}}a{-}Dur{\'{a}}n, Weston, and Yakhnenko]{bordes2013transe}
Antoine Bordes, Nicolas Usunier, Alberto Garc{\'{\i}}a{-}Dur{\'{a}}n, Jason Weston, and Oksana Yakhnenko.
\newblock Translating embeddings for modeling multi-relational data.
\newblock In \emph{{NeurIPS} 2013}, pages 2787--2795.
\newblock URL \url{https://proceedings.neurips.cc/paper/2013/hash/1cecc7a77928ca8133fa24680a88d2f9-Abstract.html}.

\bibitem[Sun et~al.()Sun, Deng, Nie, and Tang]{sun2019rotate}
Zhiqing Sun, Zhi{-}Hong Deng, Jian{-}Yun Nie, and Jian Tang.
\newblock Rotate: Knowledge graph embedding by relational rotation in complex space.
\newblock In \emph{{ICLR} 2019}. OpenReview.net.
\newblock URL \url{https://openreview.net/forum?id=HkgEQnRqYQ}.

\bibitem[Teru et~al.()Teru, Denis, and Hamilton]{teru_inductive_2020_3}
Komal Teru, Etienne Denis, and Will Hamilton.
\newblock Inductive relation prediction by subgraph reasoning.
\newblock In \emph{{ICML} 2020}, volume 119, pages 9448--9457. PMLR, 13--18 Jul .
\newblock URL \url{https://proceedings.mlr.press/v119/teru20a.html}.

\bibitem[Zha et~al.()Zha, Chen, and Yan]{zha_inductive_2021_2}
Hanwen Zha, Zhiyu Chen, and Xifeng Yan.
\newblock Inductive relation prediction by {BERT}.
\newblock In \emph{{AAAI} 2022}, pages 5923--5931. {AAAI} Press.
\newblock URL \url{https://doi.org/10.1609/aaai.v36i5.20537}.

\bibitem[Wang et~al.()Wang, Lin, and Metzler]{wang2011cascade-model}
Lidan Wang, Jimmy Lin, and Donald Metzler.
\newblock A cascade ranking model for efficient ranked retrieval.
\newblock In \emph{{SIGIR} 2011}, pages 105--114. {ACM}.
\newblock URL \url{https://doi.org/10.1145/2009916.2009934}.

\bibitem[Gal{\'{a}}rraga et~al.()Gal{\'{a}}rraga, Teflioudi, Hose, and Suchanek]{galarraga2013amie}
Luis~Antonio Gal{\'{a}}rraga, Christina Teflioudi, Katja Hose, and Fabian~M. Suchanek.
\newblock {AMIE:} association rule mining under incomplete evidence in ontological knowledge bases.
\newblock In \emph{{WWW} '13}, pages 413--422.
\newblock URL \url{https://doi.org/10.1145/2488388.2488425}.

\bibitem[Meilicke et~al.()Meilicke, Fink, Wang, Ruffinelli, Gemulla, and Stuckenschmidt]{meilicke2018rulen}
Christian Meilicke, Manuel Fink, Yanjie Wang, Daniel Ruffinelli, Rainer Gemulla, and Heiner Stuckenschmidt.
\newblock Fine-grained evaluation of rule- and embedding-based systems for knowledge graph completion.
\newblock In \emph{{ISWC} 2018}, volume 11136, pages 3--20. Springer.
\newblock URL \url{https://doi.org/10.1007/978-3-030-00671-6\_1}.

\bibitem[Yang et~al.()Yang, Yang, and Cohen]{yang2017neurallp}
Fan Yang, Zhilin Yang, and William~W. Cohen.
\newblock Differentiable learning of logical rules for knowledge base reasoning.
\newblock In \emph{{NeurIPS} 2017}, pages 2319--2328.
\newblock URL \url{https://proceedings.neurips.cc/paper/2017/hash/0e55666a4ad822e0e34299df3591d979-Abstract.html}.

\bibitem[Sadeghian et~al.()Sadeghian, Armandpour, Ding, and Wang]{sadeghian2019drum}
Ali Sadeghian, Mohammadreza Armandpour, Patrick Ding, and Daisy~Zhe Wang.
\newblock {DRUM:} end-to-end differentiable rule mining on knowledge graphs.
\newblock In \emph{{NeurIPS} 2019}, pages 15321--15331.
\newblock URL \url{https://proceedings.neurips.cc/paper/2019/hash/0c72cb7ee1512f800abe27823a792d03-Abstract.html}.

\bibitem[Yao et~al.(2019)Yao, Mao, and Luo]{yao2019kg-bert}
Liang Yao, Chengsheng Mao, and Yuan Luo.
\newblock {KG-BERT:} {BERT} for knowledge graph completion.
\newblock \emph{CoRR}, abs/1909.03193, 2019.
\newblock URL \url{http://arxiv.org/abs/1909.03193}.

\bibitem[Zimmermann(2010)]{zimmermann_fuzzy_2010}
H.-J. Zimmermann.
\newblock Fuzzy set theory.
\newblock \emph{WIREs Computational Statistics}, 2\penalty0 (3):\penalty0 317--332, 2010.
\newblock URL \url{https://wires.onlinelibrary.wiley.com/doi/abs/10.1002/wics.82}.

\bibitem[Zadeh(1965)]{zadeh_fuzzy_1965}
L.A. Zadeh.
\newblock Fuzzy sets.
\newblock \emph{Information and Control}, 8\penalty0 (3):\penalty0 338--353, 1965.
\newblock ISSN 0019-9958.
\newblock URL \url{https://www.sciencedirect.com/science/article/pii/S001999586590241X}.

\bibitem[Dettmers et~al.()Dettmers, Minervini, Stenetorp, and Riedel]{dettmers2018conve}
Tim Dettmers, Pasquale Minervini, Pontus Stenetorp, and Sebastian Riedel.
\newblock Convolutional 2d knowledge graph embeddings.
\newblock In \emph{{AAAI} 2018,}, pages 1811--1818. {AAAI} Press.
\newblock URL \url{https://doi.org/10.1609/aaai.v32i1.11573}.

\bibitem[Toutanova et~al.()Toutanova, Chen, Pantel, Poon, Choudhury, and Gamon]{toutanova2015representing}
Kristina Toutanova, Danqi Chen, Patrick Pantel, Hoifung Poon, Pallavi Choudhury, and Michael Gamon.
\newblock Representing text for joint embedding of text and knowledge bases.
\newblock In \emph{{EMNLP} 2015}, pages 1499--1509.
\newblock URL \url{https://doi.org/10.18653/v1/d15-1174}.

\bibitem[Xiong et~al.()Xiong, Hoang, and Wang]{xiong2017deeppath}
Wenhan Xiong, Thien Hoang, and William~Yang Wang.
\newblock Deeppath: {A} reinforcement learning method for knowledge graph reasoning.
\newblock In \emph{{EMNLP} 2017}, pages 564--573.
\newblock URL \url{https://doi.org/10.18653/v1/d17-1060}.

\bibitem[Su et~al.()Su, Wang, Miao, and Cui]{zhixiang2023krst}
Zhixiang Su, Di~Wang, Chunyan Miao, and Lizhen Cui.
\newblock Multi-aspect explainable inductive relation prediction by sentence transformer.
\newblock In \emph{{AAAI} 2023}, pages 6533--6540. {AAAI} Press.
\newblock URL \url{https://doi.org/10.1609/aaai.v37i5.25803}.

\end{thebibliography}
\bibliographystyle{unsrtnat}
\end{document}